\documentclass[accepted]{uai2026}
                        
\usepackage[american]{babel}
\usepackage{amssymb}

\usepackage{natbib} 
\setcitestyle{numbers,square}
\usepackage{mathtools} 
\usepackage{booktabs} 
\usepackage{tikz} 
\usetikzlibrary{automata, positioning}
\usetikzlibrary{tikzmark,fit,backgrounds}
\usepackage{subcaption}
\usepackage{amsmath}
\allowdisplaybreaks

\title{ Dictionary Based Pattern Entropy for Causal Direction Discovery}

\author[1]{\href{mailto:<harikrishnannb@goa.bits-pilani.ac.in>?Subject=Your UAI 2026 paper}{Harikrishnan N B}{}}
\author[1]{Shubham Bhilare}
\author[2]{Nithin Nagaraj}
\author[3]{Aditi Kathpalia}

\affil[1]{%
    Department of Computer Science and Information Systems\\
    BITS Pilani K K Birla Goa Campus\\
    NH 17B, Bypass, Road, Zuarinagar, 403726, Goa, India\\ harikrishnannb@goa.bits-pilani.ac.in, 2025proj022@goa.bits-pilani.ac.in
}

\affil[2]{%
    Complex Systems Programme \\
  National Institute of Advanced Studies, IISc Campus \\
  Bengaluru, 560012, Karnataka, India \\
    nithin@nias.res.in
  }
\affil[3]{%
    Department of Applied Mechanics and Biomedical Engineering\\
    Indian Institute of Technology Madras\\
    IIT P.O., Chennai 600036, India\\ aditi@iitm.ac.in
}  
  
\begin{document}
\maketitle
\begin{abstract}
Discovering causal direction from temporal observational data is particularly challenging for symbolic sequences, where functional models and noise assumptions are often unavailable. We propose a novel \emph{Dictionary Based Pattern Entropy ($DPE$)} framework that infers both the direction of causation and the specific subpatterns driving changes in the effect variable. The framework integrates \emph{Algorithmic Information Theory} (AIT) and \emph{Shannon Information Theory}. Causation is interpreted as the emergence of compact, rule based patterns in the candidate cause that systematically constrain the effect. $DPE$ constructs direction-specific dictionaries and quantifies their influence using entropy-based measures, enabling a principled link between deterministic pattern structure and stochastic variability. Causal direction is inferred via a minimum-uncertainty criterion, selecting the direction exhibiting stronger and more consistent pattern-driven organization. As summarized in Table~\ref{tab:model_comparison}, $DPE$ consistently achieves reliable performance across diverse synthetic systems, including delayed bit-flip perturbations, AR(1) coupling, 1D skew-tent maps, and sparse processes, outperforming or matching competing AIT-based methods ($ETC_E$, $ETC_P$, $LZ_P$). In biological and ecological datasets, performance is competitive, while alternative methods show advantages in specific genomic settings. Overall, the results demonstrate that minimizing pattern level uncertainty yields a robust, interpretable, and broadly applicable framework for causal discovery.
\end{abstract}

\section{Introduction}\label{sec:intro}
Modern artificial intelligence systems, particularly state of the art machine learning models, are primarily designed to estimate a functional mapping $f: \mathcal{X} \rightarrow \mathcal{Y}$ from input features to target outcomes using empirical data. These models are optimized to capture statistical regularities and associations that maximize predictive performance. However, such associations do not, in general, imply causal relationships. A learned functional dependence between variables does not establish whether a given feature exerts a causal influence on the outcome or merely correlates with it due to confounding or latent factors.

Identifying causal relationships typically requires intervention. In the standard causal inference framework, interventional queries are formalized through the $do(\cdot)$ operator. Specifically, the causal effect of $X$ on $Y$ is characterized by comparing the interventional distribution $P(Y \mid do(X=x))$ across different values of $x$. The Randomized Controlled Trial (RCT) is widely regarded as the gold standard for establishing causal effects because randomization mitigates confounding by ensuring statistical independence between treatment assignment and potential outcomes. Nevertheless, in many domains, conducting an RCT is either technically infeasible, economically impractical, or ethically impermissible. For instance, assigning individuals to smoke in order to test its effect on cancer incidence would violate fundamental ethical principles. In such settings, researchers must rely on observational data, such as historical records comparing smokers and non-smokers, to infer causal relationships.

Causal discovery from observational data aims to recover aspects of the underlying causal structure without controlled interventions. The methods used in causal discovery from observational data can be broadly classified into the following categories:
\begin{itemize}
    \item Graph Based (Constraint \& Score Based)~\cite{spirtes2000causation, chickering2002optimal}
    \item Functional Causal Models (FCMs)~\cite{shimizu2006linear, hoyer2008nonlinear}
    \item Algorithmic Information Theory (AIT)~\cite{budhathoki2016causal}
    \item Information Theoretic (Shannon-Based)~\cite{schreiber2000measuring}
    \item Temporal \& Predictive (Time Series)~\cite{granger1969investigating}
    \item Dynamical Systems (State Space)~\cite{sugihara2012detecting}

\end{itemize}
Graph-based approaches~\cite{neuberg2003causality}, rooted in structural causal models (SCMs), aim to recover causal structure by exploiting conditional independence relationships (constraint-based methods) or by optimizing global model scores over the space of possible graphs (score-based methods). Closely related but conceptually distinct are functional causal models (FCMs)~\cite{shimizu2006linear, hoyer2008nonlinear}, which identify causal direction by leveraging asymmetries in the data-generating process, such as independence between causes and noise in additive noise models or assumptions of non-Gaussianity in linear models. Both types of methods typically seek to reconstruct causal graphs or directed relationships under explicit modeling assumptions, though the assumptions and identifiability conditions differ between approaches.

In contrast, several frameworks infer causal direction without explicitly specifying structural equations. Algorithmic information-theoretic approaches build on Kolmogorov complexity~\cite{dhruthi2025causal, xu2025information, wendong2025algorithmic} and the Algorithmic Markov Condition~\cite{janzing2010causal}, asserting that the true causal direction yields a shorter description of the joint distribution than the reverse~\cite{budhathoki2016causal}. Operationally, this principle is instantiated through Minimum Description Length (MDL) formulations~\cite{grunwald2007minimum} and compression-based complexity measures.

Representative methods include ORIGO, which infers causal direction by minimizing regression-based description length~\cite{budhathoki2016causal}; ERGO, an entropy-based regression framework grounded in MDL principles~\cite{vreeken2015causal}; and Compression Complexity Causality (CCC), which leverages Effort-To-Compress (ETC) to quantify directed influence in time series~\cite{kathpalia2019data}. Related approaches employ Lempel–Ziv complexity and grammar-based compression measures to detect asymmetry via differential compressibility rather than explicit functional modeling~\cite{sy2021causal}. Collectively, these methods attribute causality to the direction that admits a more parsimonious description of the data.

Information theoretic methods quantify directed statistical dependencies or directed information flow~\cite{schreiber2000measuring}, while temporal and predictive frameworks, such as Granger causality~\cite{granger1969investigating}, infer causation from asymmetric predictability in time-ordered observations. Finally, dynamical systems approaches, including state space reconstruction and convergent cross mapping~\cite{sugihara2012detecting}, exploit properties of coupled nonlinear systems to detect causal influence from reconstructed dynamics. Together, these approaches highlight that causal discovery from observational data is not governed by a single unifying framework, but instead reflects multiple complementary perspectives on causal inference.

Despite their diversity, all approaches to causal discovery from observational data rely on assumptions that constrain identifiability. Graphical and functional models impose assumptions on conditional independence, functional form, or noise structure, while compression based and information theoretic approaches depend on approximations to algorithmic complexity or information flow. In particular, information theoretic measures require accurate probability estimation for entropy and related quantities~\cite{nagaraj2021problems}, which typically necessitates long sequences or large sample sizes. In data-scarce or highly structured observational settings, such global estimates can become unreliable, limiting their practical applicability.

In this work, we propose a methodology that combines Algorithmic Information Theory (AIT) and Shannon's Information Theory (IT) to perform causal discovery from noisy observational data.

\begin{itemize}
    \item \textbf{Algorithmic Perspective:} Inspired by AIT, we treat causal relationships as ``governing programs'' rather than statistical correlations. The method extracts a dictionary of recurring sub-patterns, which represent latent mechanistic units driving the effect. In essence, we aim to find the most compact, pattern based description of how variables influence or trigger each other.
    
    \item \textbf{Information-Theoretic Validation:} To account for noise in real data, we use Shannon entropy to measure the reliability of the extracted dictionary. Our metric, \textit{Response Determinism} ($R$-flip), quantifies how strongly a sub-pattern in $X$ predicts a change in $Y$, with values ranging from 0 (no change) to 1 (definite change). Both $0$ and $1$, indicates determinism.
\end{itemize}

Overall, the framework reveals deterministic causal structure in the data: influence values of $0$ or $1$ correspond to fully deterministic effects, implying zero entropy (no surprise or uncertainty) in the induced response. Intermediate values indicate residual uncertainty, reflecting stochastic components in the relationship. Lower uncertainty (or reduced surprise) in the direction signaling stronger deterministic influence relative to the reverse direction.

The remainder of the paper is organized as follows. Section~\ref{sec:methodolody} presents the proposed methodology, illustrated with a worked example. Section~\ref{sec:experiments} describes the experimental setup and compares the performance of the proposed approach with other AIT-based causal discovery algorithms. Section~\ref{sec:discussion} discusses the results and their implications. Section~\ref{sec:limitations} outlines the limitations of the current framework and directions for future work. Finally, Section~\ref{sec:conclusion} concludes the paper.

\section{Methodology}\label{sec:methodolody}
In this section, we provide a detailed explanation of the proposed method \emph{Dictionary Based Pattern Entropy} for causal discovery from two variables. 
\subsection{Dictionary Based Pattern Entropy (DPE)}\label{sec:problem_definition}
Let $X = (x_1, x_2, \ldots, x_N)$, $Y = (y_1, y_2, \ldots, y_N)$ denote two symbolic sequences of equal length $N$, where each symbol $x_i$ and $y_i$ takes values from a finite alphabet $\mathcal{A}$ (e.g., $\mathcal{A}=\{0,1\}$ for binary sequences). The objective of this work is to infer causal relationships between $X$ and $Y$ directly from their observed symbolic patterns, without assuming explicit probabilistic models, functional forms, or long sequence asymptotics.

We introduce a causal inference framework termed \emph{Dictionary Based Pattern Entropy ($DPE$)}, which leverages dictionary construction and pattern sensitive information measures to characterize causal asymmetries between symbolic sequences. The goals of the proposed $DPE$ framework are twofold:
\begin{enumerate}
    \item To determine the direction of causality between $X$ and $Y$ by identifying asymmetric pattern driven information changes, i.e., whether variations in $X$ induce structured changes in $Y$ or vice versa.
    \item To identify and attribute causal influence to specific patterns within the inferred causal sequence that are responsible for observable symbolic transitions in the effect sequence, thereby enabling pattern level attribution of causal strength.
\end{enumerate}

This section highlights the proposed method and a worked out example to understand the method developed to solve the problems mentioned under section~\ref{sec:problem_definition}.
\subsubsection{Causal Direction Identification}


We say that $X$ is causal for $Y$ if recurring finite patterns in $X$ 
systematically determine the transition behavior of $Y$. Under this definition, causal directionality reduces to identifying 
finite patterns in one sequence whose occurrences deterministically fix 
the transition behavior of the other. The following steps operationalize this principle to determine the 
causal direction between two variables, $X$ and $Y$. 
A worked example is provided for illustration.









Let the input sequence be a binary string $X = (x_1, x_2, \dots, x_n)$, $x_i \in \{0,1\}$. The corresponding output sequence is denoted as $Y = (y_1, y_2, \dots, y_n)$, $ y_i \in \{0,1\}$. The sequence $Y$ is constructed to reflect the occurrence of the pattern \texttt{1101} within $X$, allowing overlapping instances. That is, whenever the subsequence \texttt{1101} appears in $X$, a symbolic transition is induced in $Y$ at the corresponding position.
Formally, the output is defined as
\[
y_i =
\begin{cases}
1, & \text{if } (x_{i-3}, x_{i-2}, x_{i-1}, x_i) = (1,1,0,1),\\[6pt]
0, & \text{otherwise.}
\end{cases}
\]
Hence, the output $Y$ takes the value $1$ whenever the input sequence $X$ contains the subsequence \texttt{1101}, even if such occurrences overlap; otherwise, the output remains $0$. Consider the following symbolic sequences:
\[
X = 011101111010011001110101101001
\]
The pattern set is defined as:
\[
\text{Patterns} = \{1101\}
\]
The corresponding output sequence is:
\[
Y = 000001000010000000000100001000
\]

\begin{enumerate}
    \item \textbf{Step 1: Initialization.}  
    We define a dictionary that stores the segments (or patterns) of \( X \) corresponding to the points where a change occurs in \( Y \) as \( G_{X \rightarrow Y} \).  
    Similarly, we define \( G_{Y \rightarrow X} \) as the dictionary that stores the segments of \( Y \) whenever a change occurs in \( X \).  
    Initially, both dictionaries are empty:
    \[
    G_{X \rightarrow Y} = G_{Y \rightarrow X} = \{\}.
    \]

\item \textbf{Step 2: Construction of \( G_{X \rightarrow Y} \)}

A \emph{bit flip} in $Y$ is said to occur at position $k > 1$ if
\[
y_k \neq y_{k-1}.
\]

The dictionary $G_{X \rightarrow Y}$ is constructed by scanning $Y$ from 
left to right. Let $s$ denote the current starting index, initialized as $s=1$.

Whenever a bit flip occurs at position $k$, we extract the substring
\[
(x_s, x_{s+1}, \dots, x_k)
\]
of $X$, which explicitly includes the symbol $x_k$ aligned with the 
flip in $Y$. This substring is inserted into $G_{X \rightarrow Y}$. The starting index is then updated to $s = k+1$, and the scan continues 
until the next bit flip in $Y$. Repeated substrings are stored only once 
in the dictionary.

For the illustrative example below, consider:
\[
Y = 000001000010000000000100001000,\]
\[
X = 011101111010011001110101101001.
\]

We highlight in red the sub-pattern in \( Y \) where a bit flip occurs, and the corresponding sub-pattern in \( X \) recorded into \( G_{X \rightarrow Y} \).

\vspace{1em}
\textbf{First bit flip in \( Y \):}
    
$Y = 
\text{
\tikzmarknode{a}{\textbf{000001}}
000010000000000100001000
}$

\begin{tikzpicture}[remember picture, overlay]
    \node[draw, thick, rounded corners, red, fit=(a)] {};
\end{tikzpicture}

$X = 
\text{\tikzmarknode{b}{\textbf{011101}} 111010011001110101101001
}$

\begin{tikzpicture}[remember picture, overlay]
    \node[draw, thick, rounded corners, red, fit=(b)] {};
\end{tikzpicture}

Hence, \( G_{X \rightarrow Y} = \{011101\} \).

\vspace{1em}
\textbf{Second bit flip in \( Y \):}

$Y = 
\text{
000001 \tikzmarknode{c}{\textbf{00001}} 0000000000100001000
}$
\begin{tikzpicture}[remember picture, overlay]
    \node[draw, thick, rounded corners, red, fit=(c)] {};
\end{tikzpicture}

$X = 
\text{
011101 \tikzmarknode{d}{\textbf{11101}} 0011001110101101001
}$
\begin{tikzpicture}[remember picture, overlay]
    \node[draw, thick, rounded corners, red, fit=(d)] {};
\end{tikzpicture}

Now, \( G_{X \rightarrow Y} = \{011101,\,11101\} \).

\vspace{1em}
\textbf{Third bit flip in \( Y \):}

$Y = \text{
00000100001 \tikzmarknode{e}{\textbf{00000000001}} 00001000
}$
\begin{tikzpicture}[remember picture, overlay]
    \node[draw, thick, rounded corners, red, fit=(e)] {};
\end{tikzpicture}

$X = \text{
01110111101 \tikzmarknode{f}{\textbf{00110011101}} 01101001
}$
\begin{tikzpicture}[remember picture, overlay]
    \node[draw, thick, rounded corners, red, fit=(f)] {};
\end{tikzpicture}

So, \( G_{X \rightarrow Y} = \{011101,\,11101,\,00110011101\} \).

\vspace{1em}
\textbf{Fourth bit flip in \( Y \):}

$Y = \text{
0000010000100000000001 \tikzmarknode{g}{\textbf{00001}} 000
}$
\begin{tikzpicture}[remember picture, overlay]
    \node[draw, thick, rounded corners, red, fit=(g)] {};
\end{tikzpicture}

$X = 
\text{
0111011110100110011101 \tikzmarknode{h}{\textbf{01101}} 001
}$
\begin{tikzpicture}[remember picture, overlay]
    \node[draw, thick, rounded corners, red, fit=(h)] {};
\end{tikzpicture}

Thus, the final dictionary is:
$G_{X \rightarrow Y} = \{011101,\ 11101,\ 00110011101,\ 01101\}$.

In summary, \( G_{X \rightarrow Y} \) encodes the subpatterns in \( X \) that are temporally aligned with bit transitions in \( Y \). This dictionary can be interpreted as a directed influence mapping from \( X \) to \( Y \), capturing how variations in \( Y \) are conditioned on historical patterns of \( X \).

\item \textbf{Step 3: Construction of $G_{Y\rightarrow X}$}

As mentioned in Step 2,  whenever a bit flip occurs in \( X \), the corresponding subsequence of \( Y \) is stored in \( G_{Y \rightarrow X} \). For the example considered,
$G_{Y\rightarrow X} = \{  00, 10, 000, 10000\}$


\item \textbf{Step 4: Causal Pattern Extraction.}

For each pair of subpatterns in \( G_{X \rightarrow Y} \) and subpatterns in \( G_{Y\rightarrow X}\), we perform an \textsc{XNOR}-based sliding comparison to identify regions of strong similarity.  
Formally, given two binary sequences \( p_1 \) and \( p_2 \) of lengths \( n_1 \) and \( n_2 \) respectively (\( n_1 \leq n_2 \)), we slide \( p_1 \) across \( p_2 \) from left to right and compute the bitwise \textsc{XNOR} between overlapping bits.  
If at any position the \textsc{XNOR} result produces two or more consecutive ones, the corresponding subsequence in \( p_2 \) is extracted and stored in a \emph{pattern dictionary} as a potential causal subpattern.

Mathematically, the \textsc{XNOR} operation between two bits \( a \) and \( b \) is defined as:
\[
a \,\textsc{XNOR}\, b = 
\begin{cases}
1, & \text{if } a = b,\\
0, & \text{otherwise.}
\end{cases}
\]



\medskip
\textbf{Example:}  
Consider the two subpatterns \( p_1 = 01101 \in G_{X \rightarrow Y}\) and \( p_2 = 00110011101 \in  G_{X \rightarrow Y} \) .  
Sliding \( p_1 \) across \( p_2 \) and performing bitwise \textsc{XNOR} operations yields multiple alignments.  
At positions where two or more consecutive \textsc{XNOR} outputs are 1, we extract the corresponding subsequences from \( p_2 \) and store them in the pattern dictionary.
The below table shows that $p_1$ is sliding. 

\begin{table*}
\centering
\caption{Illustration of sliding and XNOR comparison to extract common subsequences with consecutive matches highlighted in red.}
\label{fig:slide_xnor}
\begin{subtable}[t]{0.48\textwidth}
\centering
\caption{Sliding of \( p_1 = 01101 \) over \( p_2 = 00110011101 \)}
\label{tab:slide}
\renewcommand{\arraystretch}{1.3}
\setlength{\tabcolsep}{6pt}
\begin{tabular}{c|ccccccccccc}
\toprule
 & \multicolumn{11}{c}{\textbf{Bit Positions}} \\
\midrule
\( p_2 \) & 0 & 0 & 1 & 1 & 0 & 0 & 1 & 1 & 1 & 0 & 1 \\
\midrule
\textbf{Slide 0} & 0 & 1 & 1 & 0 & 1 &   &   &   &   &   &   \\
\textbf{Slide 1} &   &\textcolor{red}{0} & \textcolor{red}{1} & \textcolor{red}{1} & \textcolor{red}{0} & 1 &   &   &   &   &   \\
\textbf{Slide 2} &   &   & 0 & 1 & 1 & \textcolor{red}{0} & \textcolor{red}{1} &   &   &   &   \\
\textbf{Slide 3} &   &   &   & 0 & 1 & 1 & 0 & 1 &   &   &   \\
\textbf{Slide 4} &   &   &   &   & 0 & 1 & 1 & 0 & 1 &   &   \\
\textbf{Slide 5} &   &   &   &   &   & \textcolor{red}{0} & \textcolor{red}{1} & \textcolor{red}{1} & 0 & 1 &   \\
\textbf{Slide 6} &   &   &   &   &   &   & 0 & \textcolor{red}{1} & \textcolor{red}{1} & \textcolor{red}{0} & \textcolor{red}{1} \\
\bottomrule
\end{tabular}
\end{subtable}
\hfill
\begin{subtable}[t]{0.47\textwidth}
\centering
\caption{XNOR results for each slide (\(1 \Rightarrow\) bit match)}
\label{tab:xnor}
\renewcommand{\arraystretch}{1.3}
\setlength{\tabcolsep}{6pt}
\begin{tabular}{c|ccccccccccc}
\toprule
 & \multicolumn{11}{c}{\textbf{Bit Positions}} \\
\midrule
\( p_2 \) & 0 & 0 & 1 & 1 & 0 & 0 & 1 & 1 & 1 & 0 & 1 \\
\midrule
\textbf{Slide 0} & 1 & 0 & 1 & 0 & 0 &   &   &   &   &   &   \\
\textbf{Slide 1} &   & \textcolor{red}{1} & \textcolor{red}{1} & \textcolor{red}{1} & \textcolor{red}{1} & 0 &   &   &   &   &   \\
\textbf{Slide 2} &   &   & 0 & 1 & 0 & \textcolor{red}{1} & \textcolor{red}{1} &   &   &   &   \\
\textbf{Slide 3} &   &   &   & 0 & 0 & 0 & 0 & 1 &   &   &   \\
\textbf{Slide 4} &   &   &   &   & 1 & 0 & 1 & 0 & 1 &   &   \\
\textbf{Slide 5} &   &   &   &   &   & \textcolor{red}{1} & \textcolor{red}{1} & \textcolor{red}{1} & 0 & 0 &   \\
\textbf{Slide 6} &   &   &   &   &   &   & 0 & \textcolor{red}{1} & \textcolor{red}{1} & \textcolor{red}{1} & \textcolor{red}{1} \\
\bottomrule
\end{tabular}

\end{subtable}

\end{table*}
From Table~\ref{tab:slide} and~\ref{tab:xnor} we identify the common subsequences between the two subpatterns 
\( p_1 = 01101 \in G_{X \rightarrow Y} \) and \( p_2 = 00110011101 \in G_{X \rightarrow Y} \)
based on consecutive \(1\)s in the XNOR comparison results.  
Let the set of all common subsequences extracted from the pair \((p_1, p_2)\) be denoted as
\(\mathcal{P}(p_1, p_2)\). Formally,
\[
\begin{aligned}
\mathcal{P}(p_1, p_2)
&= \{\, s \subseteq p_1 \mid s \text{ occurs in } p_2 \text{ whenever }\\ \text{XNOR}(p_1, p_2) &\quad  \text{ yields two or more}\\ &\quad \text{consecutive } 1\text{s} \,\}.
\end{aligned}
\]

For the given example, $\mathcal{P}(p_1, p_2) = \{0110,\; 01,\; 011,\; 1101\}$. Repeating this process for all distinct pairs of subpatterns 
\((p_i, p_j) \in G_{X \rightarrow Y} \times G_{X \rightarrow Y}\) 
with \( i \neq j \), we define the overall pattern dictionary as the union of all pairwise extractions:
\[
\mathcal{P}_{X \rightarrow Y} = \bigcup_{\substack{i,j \\ i \neq j}} \mathcal{P}(p_i, p_j).
\]
Thus, for the considered example,
$\mathcal{P}_{X \rightarrow Y} = \{01,\; 011,\; 0110,\; 011101,\; 11,\; 110,\; 1101,\; 11101\}$, and $\mathcal{P}_{Y \rightarrow X} = \{00, \; 000,\; 10\}$.

\item \textbf{Step 5: Response Determinism ($R_{\text{flip}}$).}

For each extracted common pattern, we first compute its total frequency in the candidate cause sequence (say $X$ when analyzing $X \rightarrow Y$). All occurrences are considered, including overlapping instances.

For a given pattern, we align each occurrence in $X$ with the corresponding window in $Y$ and examine whether a symbolic transition (bit flip) occurs in $Y$ within that aligned region. Let $N_{\text{occ}}$ = Total number of occurrences of the pattern in $X$,  $N_{\text{flip}}$ = Number of aligned occurrences associated with a bit flip in $Y$. The \emph{Response Determinism} of the pattern is then defined as
\[
R_{\text{flip}} = \frac{N_{\text{flip}}}{N_{\text{occ}}}.
\]

This ratio quantifies the degree to which the presence of a specific pattern in $X$ induces a symbolic transition in $Y$. Values close to $1$ indicate that the pattern consistently triggers a change in $Y$, while values close to $0$ indicate that it consistently preserves the state of $Y$. Intermediate values reflect partial or stochastic influence.


    \textbf{Example:} Consider the pattern `01'.

$X = \textcolor{red}{01}11\textcolor{red}{01}111\textcolor{red}{01}0\textcolor{red}{01}10\textcolor{red}{01}11\textcolor{red}{01}\textcolor{red}{01}1\textcolor{red}{01}0\textcolor{red}{01}.$

`$01$' occurs nine times. The corresponding windows in $Y$ are highlighted below
$Y = \textcolor{red}{00}00\textcolor{red}{01}000\textcolor{red}{01}0\textcolor{red}{00}00\textcolor{red}{00}00\textcolor{red}{0100}0\textcolor{red}{01}0\textcolor{red}{00}$. Out of nine occurrence of `01' in X, the windows in $Y$ that changed are only four, the windows that didn't change are 5. $R_{flip}$ for $01$ is $\frac{4}{9}$. The below table (Table~\ref{table_pattern_frequency_x_y}) computes all $R_{flip}$ for all the common patterns in $\mathcal{P}_{X \rightarrow Y}$. 
\begin{table}[h!]
\centering
\caption{Pattern statistics showing the number of changes, no-changes, and the ratio (From $X \rightarrow Y$). }
\label{table_pattern_frequency_x_y}
\begin{tabular}{|c|c|c|c|}
\hline
\textbf{Pattern} & \textbf{Change Count} & \textbf{No Change Count} & \textbf{Ratio} \\
\hline
01     & 4 & 5 & 0.444 \\
011    & 1 & 4 & 0.200 \\
0110   & 0 & 2 & 0.000 \\
011101 & 2 & 0 & 1.000\\
11     & 1 & 8 & 0.111 \\
110 & 0 & 5 & 0.000\\
1101   & 4 & 0 & 1.000 \\
11101  & 3 & 0 & 1.000 \\
\hline
\end{tabular}

\end{table}

Similarly, we compute $R_{flip}$ for all the common patterns in $\mathcal{P}_{Y \rightarrow X}$ (Table~\ref{table_pattern_frequency_y_x}). 
\begin{table}[h!]
\centering
\caption{Pattern statistics showing the number of changes, no-changes, and the ratio (From $Y \rightarrow X$).}
\label{table_pattern_frequency_y_x}
\begin{tabular}{|c|c|c|c|}
\hline
\textbf{Pattern} & \textbf{Change Count} & \textbf{No Change Count} & \textbf{Ratio} \\
\hline
00     & 10 & 11 & 0.476 \\
000    & 13 & 3 & 0.8125 \\
10   & 3 & 1 & 0.75 \\

\hline
\end{tabular}

\end{table}




 \item \textbf{Step 6: Calculating weighted entropy.}\\

The causal relationship between two binary sequences \( X \) and \( Y \) is inferred using the principle of \emph{Minimum Uncertainty}. The central idea is that the true causal direction should exhibit more deterministic pattern-level influence and therefore lower uncertainty.

For each extracted pattern \( p \in \mathcal{P} \), we compute a \emph{Weighted Binary Entropy} that quantifies the uncertainty associated with the influence of \( p \).

The weighted entropy of a pattern \( p \) is defined as
\begin{equation}
H_w(p) = W_p \, H_b(r_p) \quad \text{bits},
\end{equation}
where \( r_p \) (the $R_{\text{flip}}$ value of pattern \( p \)) denotes the flip contribution ratio of pattern \( p \), and \( H_b(\cdot) \) is the binary entropy function
\begin{equation}
H_b(r_p) = -\bigl[ r_p \log_2 r_p + (1 - r_p)\log_2(1 - r_p) \bigr].
\end{equation}

The weight \( W_p \) represents the normalized frequency of occurrence of pattern \( p \) and is defined as
\begin{equation}
W_p = \frac{C_p}{N - L + 1},
\end{equation}
where \( C_p \) is the total number of occurrences of pattern \( p \) in the assumed causal sequence, \( N \) is the sequence length, and \( L \) is the length of the pattern. The quantity \( N - L + 1 \) denotes the total number of possible substrings of length \( L \) in the assumed causal sequence.

To evaluate an overall direction, we compute the \emph{Average Weighted Entropy} across the pattern set \( \mathcal{P} \):
\begin{equation}
\bar{H}
=
\frac{1}{|\mathcal{P}|}
\sum_{p \in \mathcal{P}} H_w(p)
\quad \text{bits}.
\end{equation}

Causal direction is determined by comparing the aggregate uncertainties:
\[
\bar{H}_{X \rightarrow Y}
\quad \text{and} \quad
\bar{H}_{Y \rightarrow X}.
\]
The direction yielding the smaller value of \( \bar{H} \) is inferred as causal, as it corresponds to lower pattern-level uncertainty and stronger deterministic structure.

\item \textbf{Step 7: Directional Comparison and Causal Verdict} \\
The direction yielding the lower value of \( \bar{H} \) corresponds to \emph{minimum uncertainty} and thus reflects greater confidence in the predictable influence of the causal variable on the effect variable. This direction is therefore inferred as the causal direction.

         \begin{equation*}
        \text{Verdict} = 
        \begin{cases} 
              X \to Y & \text{if } \bar{H}_{X \to Y} < \bar{H}_{Y \to X} \\
              Y \to X & \text{if } \bar{H}_{X \to Y} > \bar{H}_{Y \to X} \\
              \text{Independent} & \text{if } \bar{H}_{X \to Y} = \bar{H}_{Y \to X}
        \end{cases}
        \end{equation*}
       
\end{enumerate}

We illustrate the computation involved in Step 7 and step 8 using the below worked out example. Let us consider the symbolic sequence of X and Y mentioned in the earlier example.

        \[
        Y = 000001000010000000000100001000,\]
        \[X = 011101111010011001110101101001.
        \]
        We use the formula: \begin{equation*}
    H_w(p) = W_p \cdot \left( -\left[ r_p \log_2(r_p) + (1 - r_p) \log_2(1 - r_p) \right] \right)
    \end{equation*}

To illustrate the methodology, we compute $H_b(r_p)$, $H_w(p)$, and $\bar{H}_{X \rightarrow Y}$ assuming $X$ as the causal variable and $Y$ as the effect variable. The same steps can be followed for the computation of $\bar{H}_{Y \rightarrow X}$. Consider the pattern $p = 01 \in \mathcal{P}_{X \rightarrow Y}$.

\begin{figure}[!ht]
    \centering
    \includegraphics[width=0.50\textwidth]{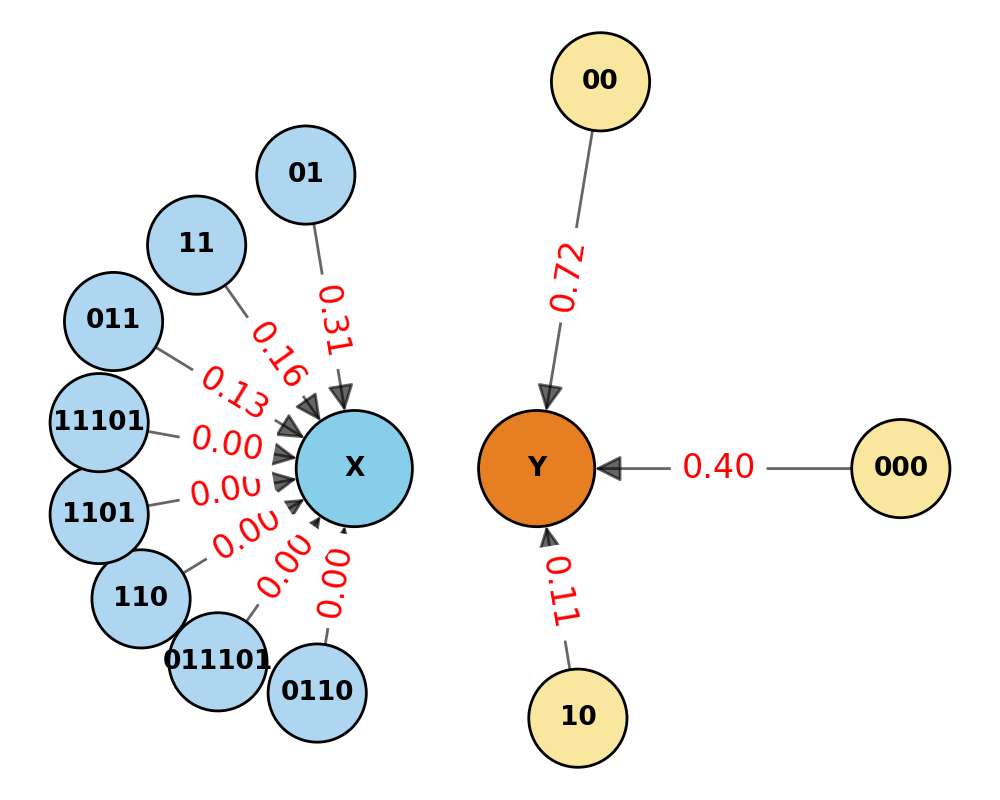}
    \caption{Directed network of symbolic patterns illustrating deterministic structure 
between $X$ and $Y$. Nodes represent distinct substrings extracted from the 
driving sequence, and edges encode their weighted entropy with 
respect to the target sequence. A weighted entropy value close to $0$ 
indicates that the pattern induces highly deterministic transition behavior 
in the target, whereas larger values indicate increasing uncertainty in the 
induced transitions. The figure displays both directions, $X \rightarrow Y$ 
and $Y \rightarrow X$, enabling comparison of directional determinism.}
    \label{fig:36}
\end{figure}

    \begin{enumerate}
        \item \textbf{Binary Entropy Calculation ($H_b(p)$):} 
        For pattern $p = 01$, the ratio $r_p = 0.444$ (Table~\ref{table_pattern_frequency_x_y}).
        $ H_b(0.444)  \approx 0.991 $ \text{ bits}
    
        \item \textbf{Frequency Weight Calculation ($W_p$):}
        With count $C_p = 9$, total length $N = 30$, and pattern length $L = 2$:
        \[ W_p = \frac{9}{30 - 2 + 1} = \frac{9}{29} \approx 0.310 \]
    
        \item \textbf{Weighted Entropy ($H_w$):}
        \[ H_w(01) = W_p \cdot H_b(r) = 0.310 \cdot 0.991 = 0.307 \text{ bits}\]
        The same computation is performed for all remaining patterns in \( \mathcal{P}_{X \rightarrow Y} \).

        \item \textbf{Average Weighted Entropy ($\bar{H}$):}
        The total Average Weighted Entropy for the set $P_{X \to Y}$ is the sum of all individual $H_w$ values divided by the total number of unique patterns ($|P_{X \to Y}| = 8$):
        \[ \bar{H}_{X \to Y} = \frac{1}{8} \sum_{p \in P} H_w(p) \]
        \[ \bar{H}_{X \to Y} = \frac{0.307 + 0.129 + 0 + 0 + 0.156 + 0 + 0 + 0}{8}  \]
          \[ \bar{H}_{X \to Y} = 0.074 \text{ bits}\]

        The same procedure can be followed to compute $\bar{H}_{Y \to X}$.

        \begin{table*}
    \centering
    \caption{Entropy and Weights for Patterns (From $X \rightarrow Y$)}
    \label{table_entropy_x_y}
    \begin{tabular}{|c|c|c|c|c|c|}
    \hline
    \textbf{Pattern ($p$)} & \textbf{Ratio ($r$)} & \textbf{Weight ($W_p$)} & \textbf{Binary Entropy $H_b(r_p)$ (bits)} & \textbf{Weighted Entropy $H_w(p)$ (bits)} \\
    \hline
    01     & 0.444 & 0.310 & 0.991 & 0.307 \\
    011    & 0.200 & 0.179 & 0.722 & 0.129 \\
    0110   & 0.000 & 0.074 & 0.000 & 0.000 \\
    011101 & 1.000 & 0.080 & 0.000 & 0.000 \\
    11     & 0.111 & 0.310 & 0.503 & 0.156 \\
    110    & 0.000 & 0.179 & 0.000 & 0.000 \\
    1101   & 1.000 & 0.148 & 0.000 & 0.000 \\
    11101  & 1.000 & 0.115 & 0.000 & 0.000 \\
    \hline
    \multicolumn{4}{|r|}{\textbf{Average Weighted Entropy $\bar{H}_{X \rightarrow Y}$}} & \textbf{0.074} \\
    \hline
    \end{tabular}
    \end{table*}

    \begin{table*}
    \centering
    \caption{Entropy and Weights for Patterns (From $Y \rightarrow X$)}
    \label{table_entropy_y_x}
    \begin{tabular}{|c|c|c|c|c|c|}
    \hline
    \textbf{Pattern ($p$)} & \textbf{Ratio ($r$)} & \textbf{Weight ($W_p$)} & \textbf{Binary Entropy $H_b(r_p)$ (bits)} & \textbf{Weighted Entropy $H_w(p)$ (bits)} \\
    \hline
    00  & 0.476  & 0.724 & 0.999 & 0.723 \\
    000 & 0.8125 & 0.571 & 0.696 & 0.397 \\
    10  & 0.750  & 0.138 & 0.811 & 0.112 \\
    \hline
    \multicolumn{4}{|r|}{\textbf{Average Weighted Entropy $\bar{H}_{Y \rightarrow X}$}} & \textbf{0.411} \\
    \hline
    \end{tabular}
    \end{table*}
      \item \textbf{Causal Verdict}
    For the example considered, $\bar{H}_{X \rightarrow Y} = 0.074$ and $\bar{H}_{Y \rightarrow X} = 0.411$, we observe:
    \[ \bar{H}_{X \rightarrow Y} < \bar{H}_{Y \rightarrow X} \]
    Therefore, the causal direction is identified as \textbf{$X \rightarrow Y$}. 
\end{enumerate}
Figure~\ref{fig:36} visualizes the pattern transition structure as a 
directed network. Each node corresponds to a unique substring extracted 
from the source sequence, and the associated weighted entropy quantifies 
the uncertainty of the induced transition behavior in the target sequence. 
Lower entropy values indicate deterministic control, while higher values 
reflect variability in the induced transitions. The dual construction 
($X \rightarrow Y$ and $Y \rightarrow X$) allows assessment of asymmetry 
in deterministic structure, which forms the basis for causal direction 
identification.
    
\section{Experiments and Results}\label{sec:experiments}
In this section, we discuss the experimental setup and comparative performance analysis of the $DPE$ causal framework with existing methods based on recent algorithmic information theory in the literature, such as Lempel-Ziv penalty ($LZ_P)$~\cite{sy2021causal},  Effort-To Compress- Efficacy ($ETC_E$)~\cite{sy2021causal}, Effort-To-Compress Penalty ($ETC_P$)~\cite{sy2021causal}. 
\subsection{Effect of Delayed Bit-flip for a pattern on Direction of Causality:}\label{sec:experiment8}
Consider a sequence $X$ with the pattern $1101$ as the most repeating pattern and $Y$ be the sequence with a delayed bit flip after the pattern $1101$ occurs in $X$. Both sequences are of length 100. We generated 1000 pairs of sequences for each delay, varied from 0 to 6. The direction of causality ideally should be from $X\rightarrow Y$.\\

\begin{itemize}
    \item \textbf{Causality:} $X \rightarrow Y$ \\
    \item \textbf{Pattern:} $1101$ \\
   \item  \textbf{Delay:} $k = 2$ \\
\end{itemize}

$$
\begin{aligned}
X &= \dots x, \mathbf{1, 1, 0, 1}, x_{i+1}, x_{i+2}, \dots \\
Y &= \dots y, 0, 0, 0, 0, 0, \mathbf{1},y_{i+3} \dots
\end{aligned}
$$

\text{Example with bitstrings } $x, y \in \{0, 1\}^*$:
$$
\begin{aligned}
X &= \dots \mathbf{1101} \dots \\
Y &= \dots 0000 \mathbf{0}\mathbf{0}1 \dots
\end{aligned}
$$
\subsubsection{Results: Effect of Delayed Bit-flip for a pattern on Direction of Causality:}
\begin{figure}[!ht]
    \centering
    \includegraphics[width=0.50\textwidth]{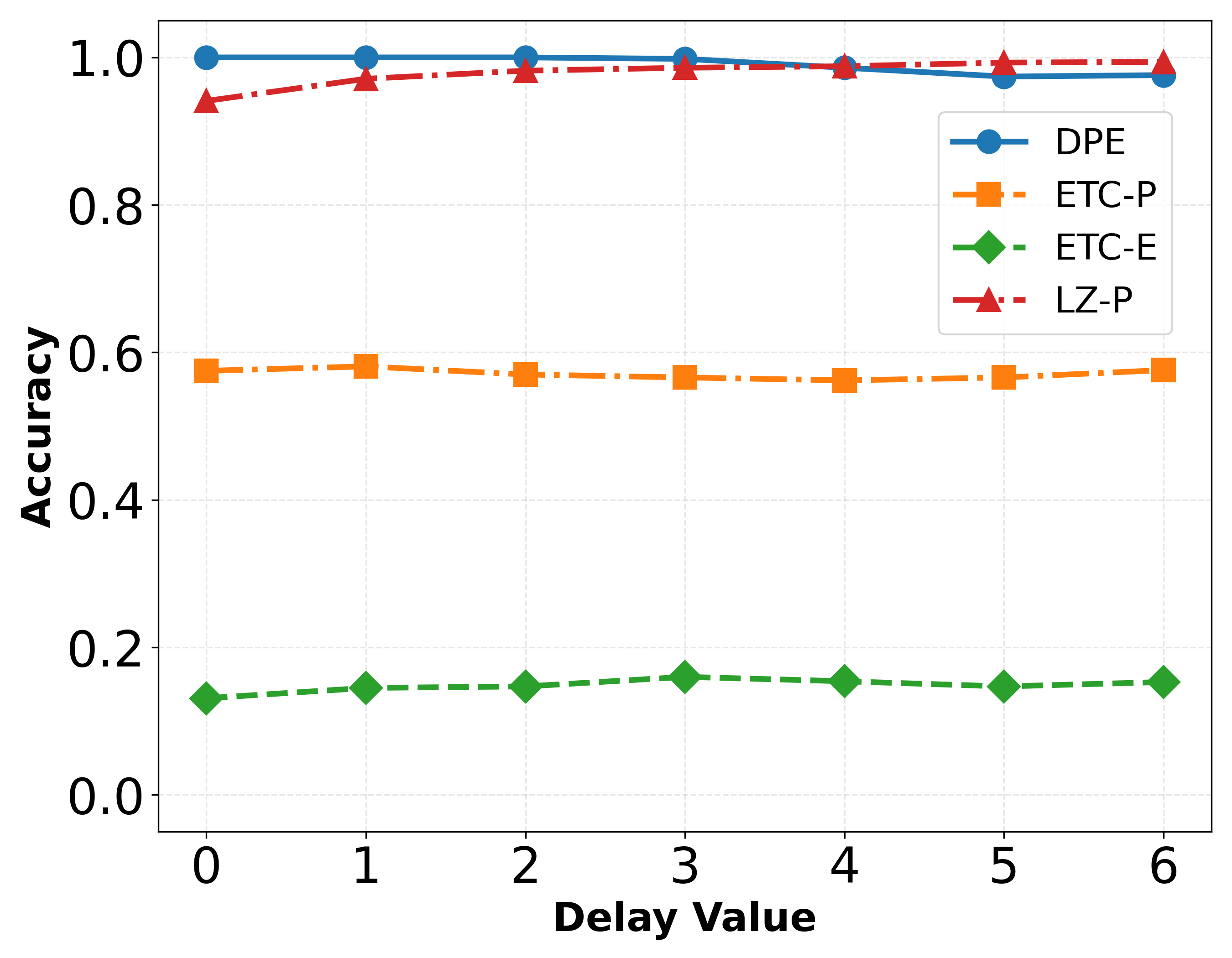}
\caption{Effect of delayed bit-flip on causal direction detection: accuracy versus bit-flip delay for $DPE$, $ETC_P$, $ETC_E$, and $LZ_P$.}
    \label{fig:31}
\end{figure}

As mentioned in Experiment \ref{sec:experiment8}, the ground truth for this experiment is $X\rightarrow Y$ (X causes Y). The $DPE$ correctly predicts the direction of causality with an average accuracy of $99\%$ across $k$ from $0$ to $6$, $LZ_p$ shows similar results, accurately predicting $97.9\%$ pairs correctly can be clearly seen in Figure\ref{fig:31}. While the $ETC_P$ was able to predict direction of causality correctly in $57\%$ pairs only, and $ETC_E$ completely fails in such cases.

\subsection{Synthetic unidirectional coupling}\label{sec:experiment1}
The autoregressive (AR) model is widely used to describe time-varying processes that linearly depend on their past and the past of the other processes \cite{shumway2017time}. We used the AR model for unidirectional causal inference and simulated autoregressive processes of order one (AR(1)) as follows, with $X_t$ and $Y_t$ as the dependent and independent processes, respectively.

\begin{equation*}
    X_t = a X_{t-1} + \eta Y_{t-1} + \epsilon_{x,t}
\end{equation*}
\begin{equation*}
    Y_t = b Y_{t-1} + \epsilon_{y,t}
\end{equation*}

The coefficients of $X_t$ and $Y_t$ were fixed as \( a = b = 0.8 \). We generated length time series for $X_t$ and $Y_t$. The coupling strength \( \phi \) was varied from \( 0 \) to \( 0.95 \) in steps of \( 0.05 \), and additive noise with intensity \( \nu = 0.01 \) drawn from a standard normal distribution was included. For each value of $\phi$, we generated $2,000$ independent trials of $X_t$, $Y_t$ pairs of length $1500$, and first $500$ transient values were removed.  Each discrete time was encoded into binary sequences using an equi-width binning strategy. Causal directionality and strengths were subsequently estimated across all four models for every trial, resulting in a comprehensive evaluation over $40,000$ total trials.
\subsubsection{Results: AR Process Synthetic unidirectional coupling:}\label{sec:results1}

For $DPE$ causal discovery framework, for $\phi=0.05$ and $0.20$, the model starts to perform better, with accuracy rising quickly from $55.15\%$ to $91.95\%$. From $\phi= 0.2$ to $0.35$, the model reliability increases, with accuracy over $98.5\%$. Finally, when the coupling strength is $0.40$ or higher, the model reaches almost perfect accuracy, between 99\% and 100\%. $DPE$ outperforms $ETC_E$ and $ETC_P$ for all values of $\phi$ greater than $0$. $DPE$ has a performance similar to that of $LZ_P$.
\begin{figure*}
    \centering
    \begin{minipage}{0.50\textwidth}
     \centering
    \includegraphics[width=\textwidth]{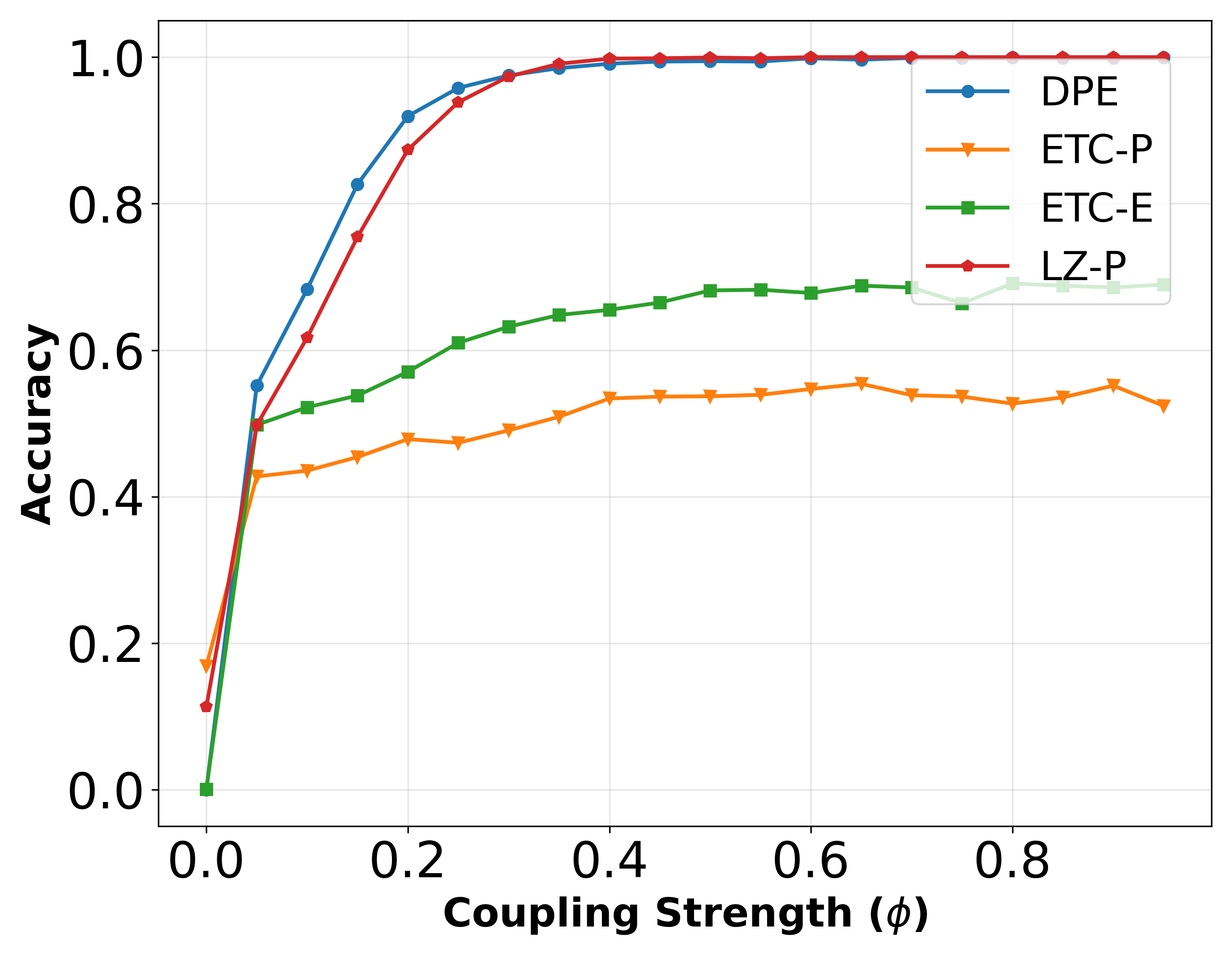}
\caption{Synthetic unidirectional coupling: accuracy of $DPE$, $ETC_E$, $ETC_P$, and $LZ_P$ versus coupling strength.}
    \label{fig:24}
     \end{minipage}
    \hfill 
    \begin{minipage}{0.45\textwidth}
        \centering
        \includegraphics[width=\textwidth]{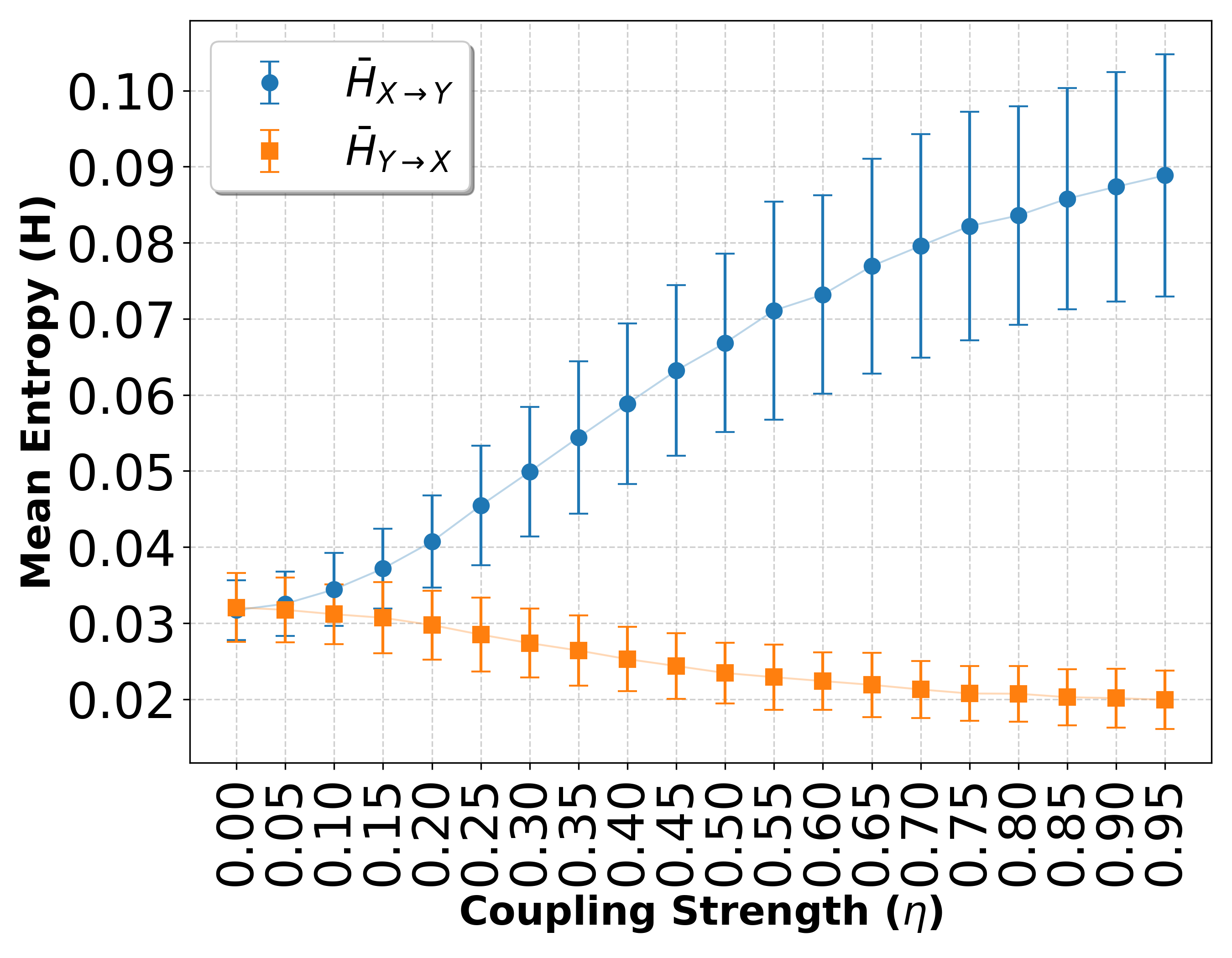}
        \caption{Synthetic unidirectional coupling: average entropy ($\bar{H}_{X \to Y}$ and $\bar{H}_{Y \to X}$) versus coupling strength ($\phi$) for the $DPE$ causal discovery framework. The increasing separation between the two curves indicates improved discrimination between the driving and response variables.}
        \label{fig:19}
    \end{minipage}
\end{figure*}

     
This performance is further explained by the behavior of the directional entropy measures shown in Figure \ref{fig:19}, which plots the average entropy $\bar{H}_{X \to Y}$ and $\bar{H}_{Y \to X}$ against coupling strength. The divergence between these two entropy values confirms the model's increasing ability to identify the true causal driver as the coupling strength intensifies.
\subsection{Sparse Processes}

Let $T = \{1,2,\dots,n\}$ denote the set of discrete time indices, where $n=2000$. Let $Z_1(t)$ and $Z_2(t)$ be two stochastic processes defined for $t \in T$. Let $T_1 \subseteq T$ be a subset such that $|T_1| = k$, where the elements of $T_1$ are selected uniformly at random without replacement from $T$. Define $T_2 = \{\, t+1 \mid t \in T_1,\; t+1 \in T \,\}$. Thus, $T_2$ consists of the immediate successor time indices of elements in $T_1$, restricted to remain within $T$. Let $T_1^c = T \setminus T_1$ and $T_2^c = T \setminus T_2$ denote the complements of $T_1$ and $T_2$, respectively. The latent processes are defined as

\[
Z_1(t) = \alpha Z_1(t-1) + \epsilon_1(t),
\]

\[
Z_2(t) = \beta Z_2(t-1) + \gamma z_1(t-1) + \epsilon_2(t),
\]

where $\epsilon_1(t)$ and $\epsilon_2(t)$ are independent Gaussian noise terms drawn from $ \mathcal{N}(0, 0.1)$. The parameters are fixed as $\alpha = 0.8, \quad \beta = 0.08, \quad \gamma = 0.75$. The observed sparse sequences $z_1(t)$ and $z_2(t)$ are defined as

\[
z_1(t) =
\begin{cases}
Z_1(t), & t \in T_1, \\
0, & t \in T_1^c,
\end{cases}
\]

\[
z_2(t) =
\begin{cases}
Z_2(t), & t \in T_2, \\
0, & t \in T_2^c.
\end{cases}
\]

For analysis, the sparse sequences are binarized as follows:

\[
z_1^{(b)}(t) =
\begin{cases}
1, & z_1(t) \neq 0, \\
0, & z_1(t) = 0,
\end{cases}
\]

\[
z_2^{(b)}(t) =
\begin{cases}
1, & z_2(t) \neq 0, \\
0, & z_2(t) = 0.
\end{cases}
\]

The sparsity parameter $k$ is varied from $5$ to $50$ in increments of $5$.  
For each value of $k$, $100$ independent trials are performed.

\subsubsection{Results: Sparse Processes}

\begin{figure}[!ht]
    \centering
    \includegraphics[width=0.50\textwidth]{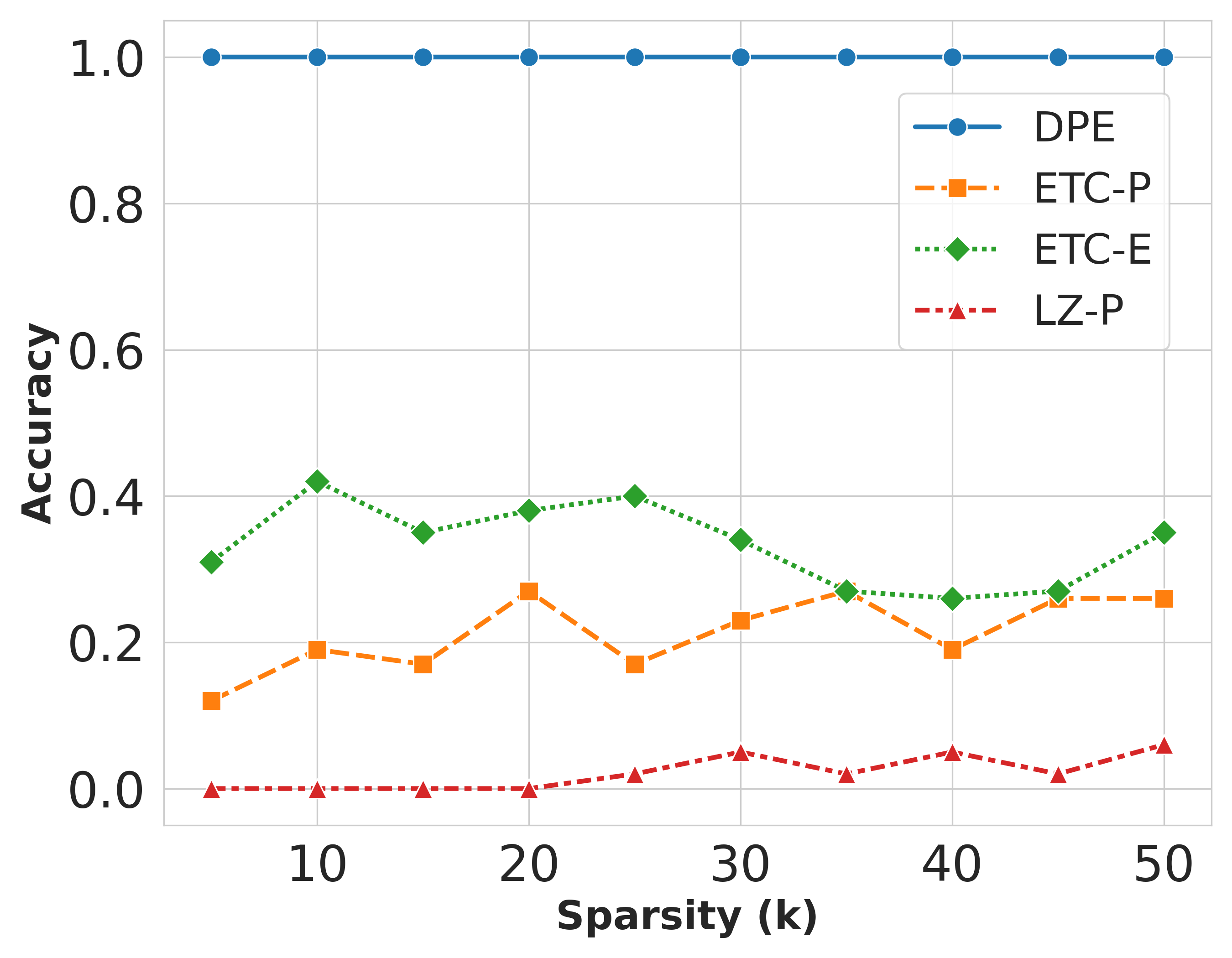}
\caption{Sparse processes: comparison of accuracy for $DPE$, $ETC_E$, $ETC_P$, and $LZ_P$ across varying sparsity levels ($k$).}
    \label{fig:34}
\end{figure}

For binary discretizations of the sparse data the performance of the $DPE$ is 100\% accurate for all values of sparsity as depicted in Figure \ref{fig:34}, while $LZ_P$, $ETC_E$ and $ETC_P$ shows decrease in performance predicting most proportions of sequences as independent. 



\subsection{Coupled 1D Skew-Tent Maps}\label{sec:experiment2}

To evaluate the framework's effectiveness on non-linear timeseries data, we utilize a coupled driver-response configuration of 1D skew-tent. The map function $T(x, b)$ is a piecewise linear chaotic map defined as:

\begin{equation*}
    T(x, b) = 
    \begin{cases} 
    x/b & \text{if } 0 \leq x < b \\
    (1-x)/(1-b) & \text{if } b \leq x < 1 
    \end{cases}
\end{equation*}

The governing equations used to generate the driver ($D_t$) and response ($R_t$) time series follow a unidirectional coupling scheme:

\begin{equation*}
    D_t = T_1(D_{t-1}, b_1)
\end{equation*}
\begin{equation*}
    R_t = (1 - \eta) T_2(R_{t - 1}, b_2) + \eta D_t
\end{equation*}

\noindent where $D_t$ represents the driver (cause) and $R_t$ represents the response (effect). Note that the response system’s state at time $t$ is influenced by the master’s concurrent state $D_t$. The skewness parameters of the maps are fixed at \( b_1 = 0.35 \) for the driving system and \( b_2 = 0.76 \) for the response system. The coupling coefficient \( \eta \) is varied from \( 0.0 \) to \( 0.9 \) in increments of \( 0.1 \). For each value of \( \eta \), \( 2000 \) independent trials with each  trial generating $1500$ length $D_t$ and $S_t$ timeseries data. We remove the first  $500$ transient values. Initial conditions for each trial are drawn from a uniform distribution \( U(0,1) \).

\subsubsection{Results: Coupled 1D Skew-Tent Map:}\label{sec:results2}

$DPE$ detects the correct direction of causation for coupling strengths ($\eta$) greater than $0$ outperforming all the models with overall accuracy of $90\%$. While, $LZ_P$ shows an overall accuracy of $62.84\%$ performing better than $ETC_E$ ($22.48\%$) and $ETC_P$ ($49.88\%$). From a timeseries point of view, a coupling coefficient of $0.9$ leads to synchronization of timeseries leading many methods to fail in detecting causality, but $DPE$ detects the direction of causation with $100\%$ accuracy.

\begin{figure*}
    \centering
    \begin{minipage}{0.5\textwidth}
        \centering
        \includegraphics[width=\textwidth]{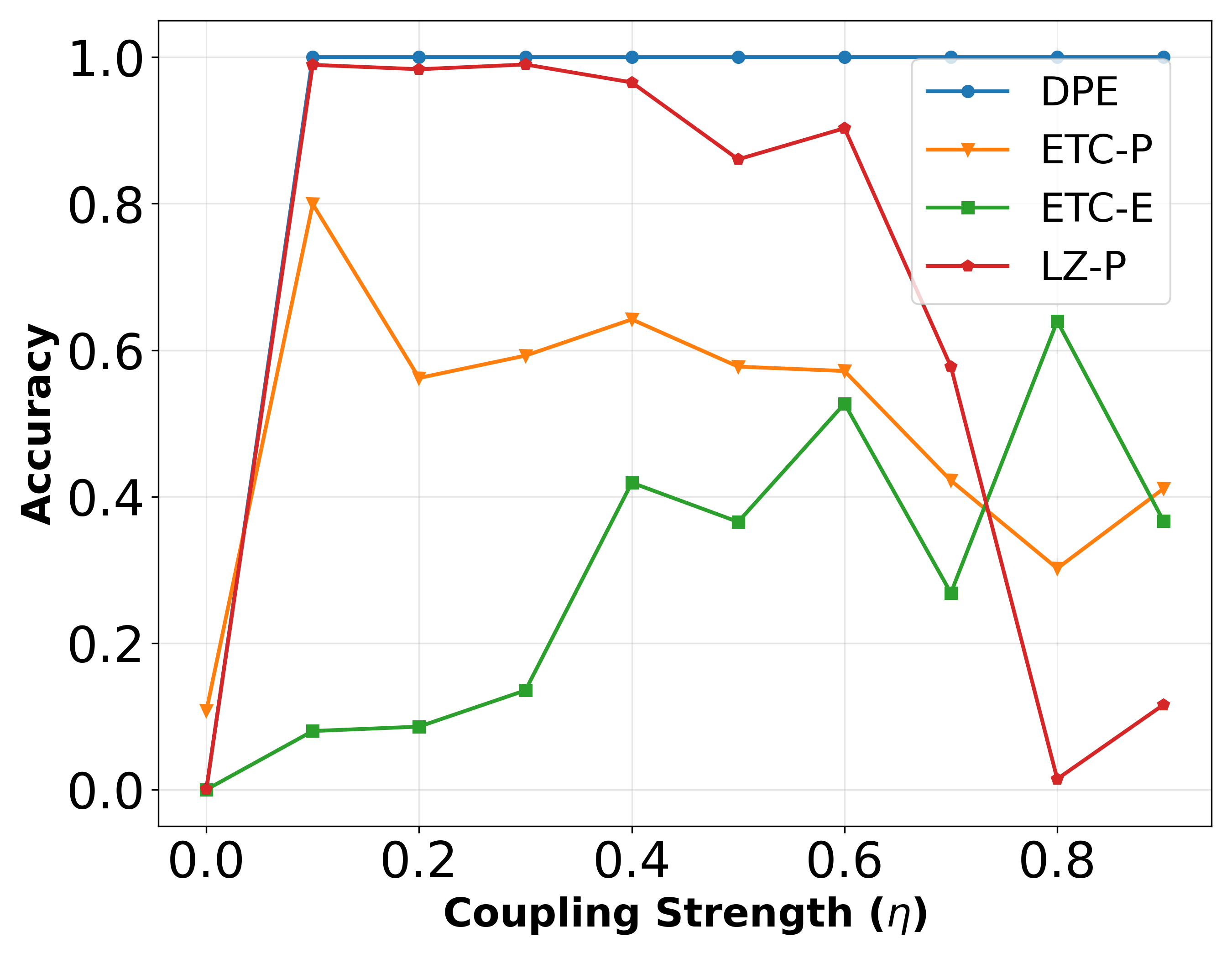}
  \caption{Coupled 1D skew-tent maps: accuracy of $DPE$, $ETC_E$, $ETC_P$, and $LZ_P$ across varying coupling strengths.}
    \label{fig:23}
    \end{minipage}
    \hfill 
    \begin{minipage}{0.45\textwidth}
        \centering
        \includegraphics[width=\textwidth]{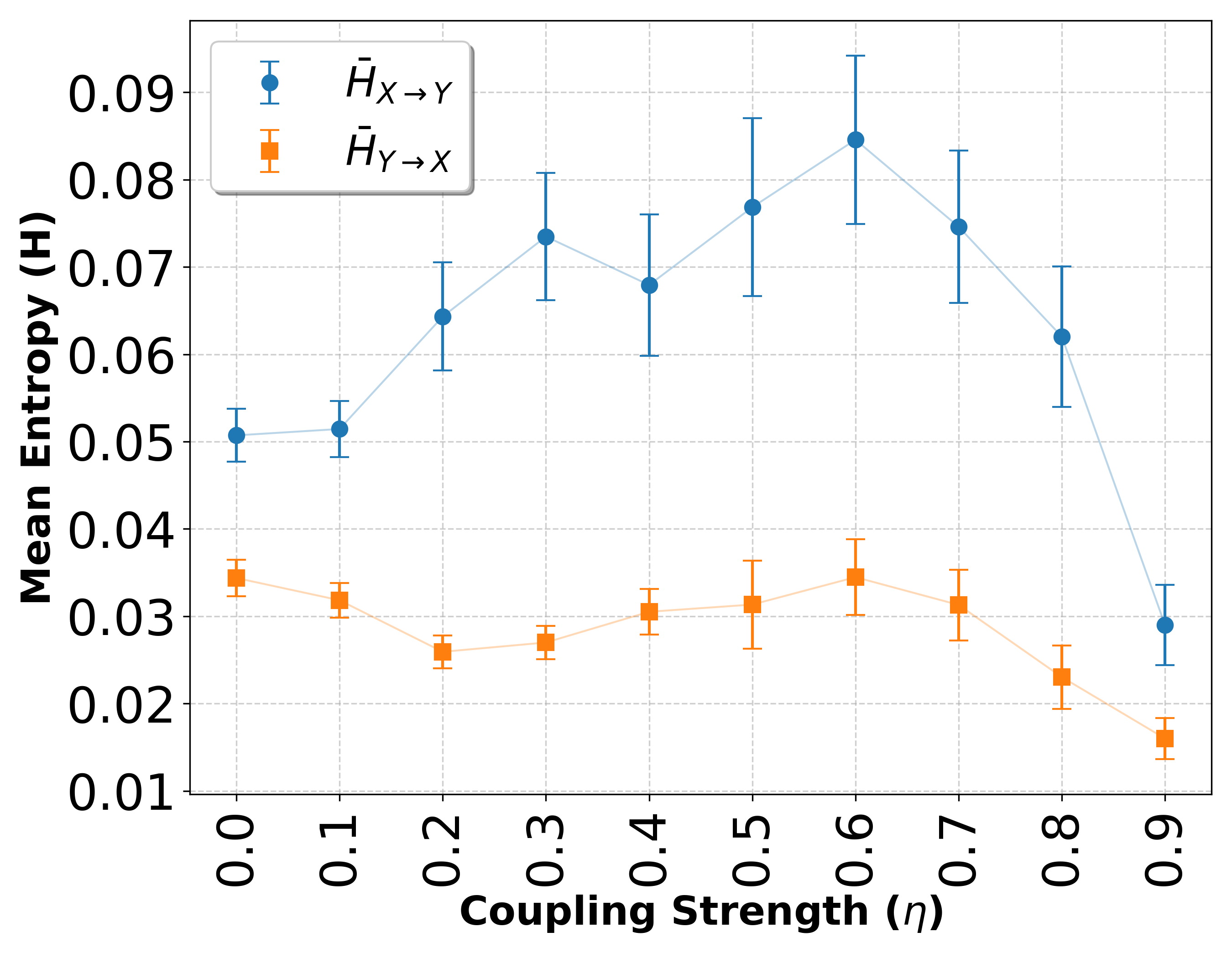}
      \caption{Coupled 1D skew-tent maps: Average entropy values ($\bar{H}_{X \to Y}$ and $\bar{H}_{Y \to X}$) obtained using the DPE causal discovery framework for the coupled 1D skew-tent map experiment.}
        \label{fig:21}
    \end{minipage}
\end{figure*}



\subsection{Genomic Causal Analysis of SARS-CoV-2}\label{sec:experiment3}


This experiment assesses whether the country-specific initial sequence (CW) or the global Reference Sequence (RS) SARS-CoV-2 (GenBank
Accession ID: $NC\_045512.2$) \cite{coronaviridae2020species} serves as the main cause for the later domestic viral evolution. We test the null hypothesis ($H_0$) that the global consensus sequence (RS) causes all subsequent domestic sequences against the alternative ($H_1$) that the first sequence reported within a particular country (CW) exerts a stronger influence\cite{sy2021causal} using a dataset of 17,567 high-quality nucleotide sequences from 19 countries in GenBank Database \cite{genbank_website}. Numerical mapping (A=1, C=2, G=3, T=4) is used to evaluate genomic sequences from a variety of countries, such as China, France, and India, across random subsets.

\subsubsection{Results: Genomic Causal Analysis of SARS-CoV-2:}\label{sec:results3}

\begin{figure}[!ht]
    \centering
    \includegraphics[width=0.50\textwidth]{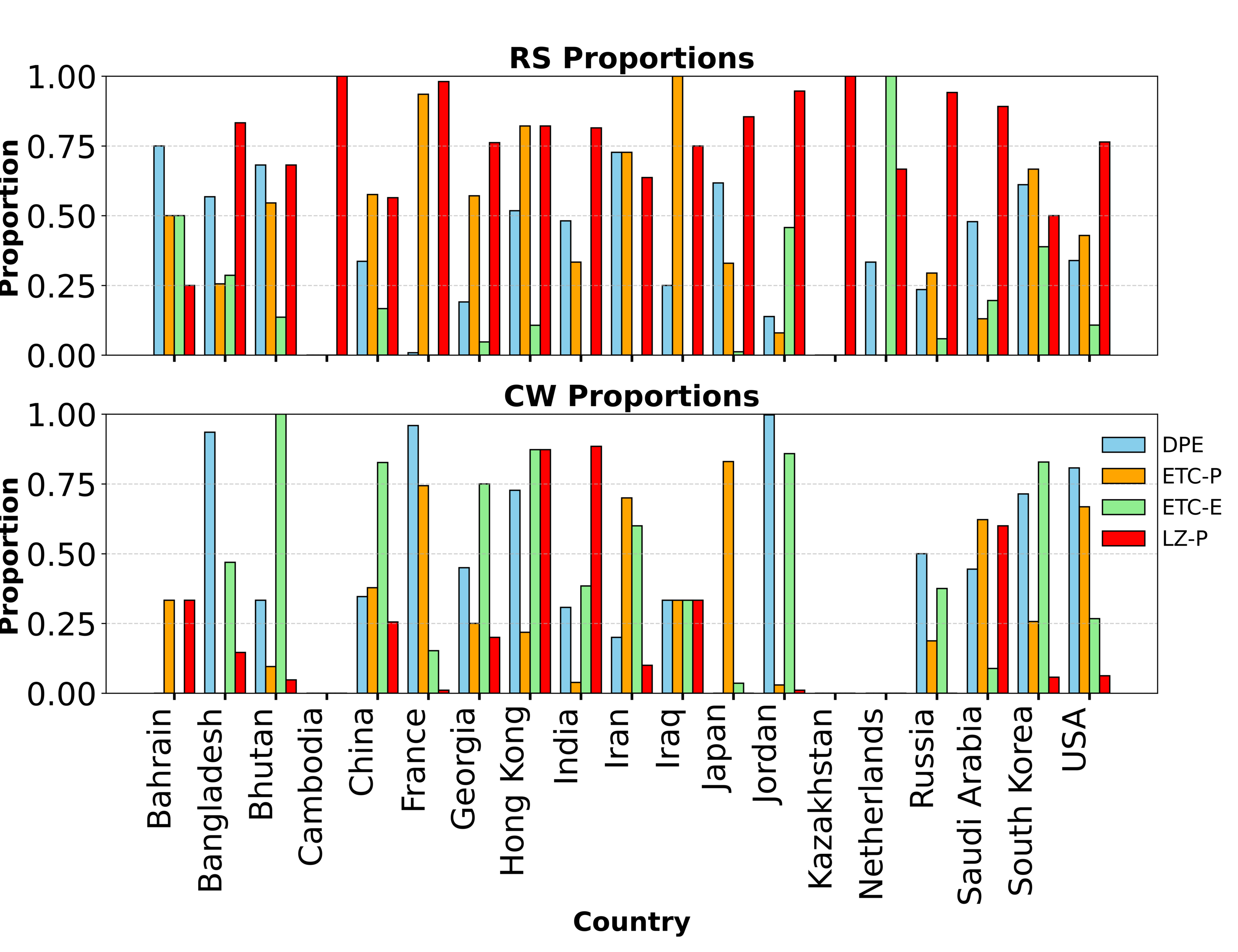}
    \caption{Genomic causal analysis of SARS-CoV-2: Comparative summary of $DPE$, $ETC_E$, $ETC_P$, and $LZ_P$ sensitivity to global (RS) vs. local (CW) evolutionary drivers.}
    \label{fig:25}
\end{figure}

We analyzed whether the SARS-CoV-2 consensus sequence (RS) `causes' all other sequences. 16 countries for both $DPE$ and $ETC_P$, 11 countries for $ETC_E$ and 19 for $LZ_P$ out 19 countries had atleast $5\%$ sequences which admitted the $H_0$ hypothesis direction. The Figure \ref{fig:25} illustrates that the higher proportions across countries were generally observed under $ETC_P$ and $LZ_P$ models .

The proportions whether the sequences per country `caused' by CW were greater than RS. $DPE$ shows 10 countries, 14 countries for $ETC_E$ and 3 countries for both $ETC_P$ and $LZ_P$ out of 19 countries. The proportions per country can be seen in the Figure \ref{fig:25}.

\subsection{Predator-Prey System}
\begin{figure}[!ht]
    \centering
    \includegraphics[width=0.50\textwidth]{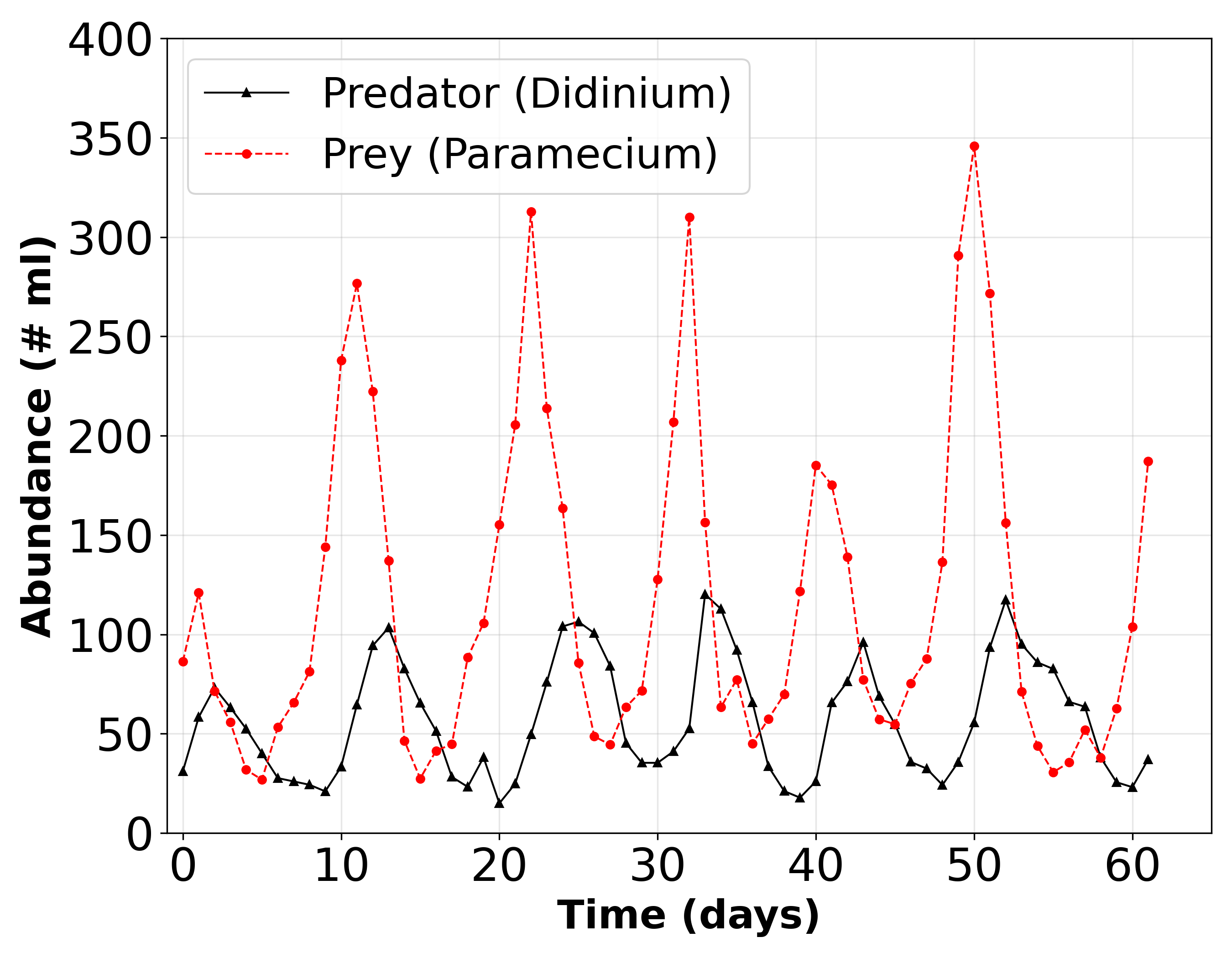}
    \caption{Population dynamics of the predator (Didinium nasutum) and prey (Paramecium aurelia).}
    \label{fig:predator-prey}
\end{figure}

$DPE$ was evaluated on a real-world dataset from a prey-predator system \cite{veilleux1976analysis}\cite{jost2000testing}. The data consists of 71 data points of predator (Didinium nasutum) and prey (Paramecium aurelia) populations (Figure \ref{fig:predator-prey}. The predator population influences the prey population directly, and then itself gets influenced by a change in the prey population, indicating a system of bidirectional causation. The direction of causal influence from predator to prey is expected to be higher than in the opposite direction. After removing the initial 9 transients, the remaining 62 data points were considered for this analysis.

\subsubsection{Results: Predator-Prey System}
\begin{table}[htbp]
\centering
\caption{Predator-Prey System: Comparative causal metrics for Predator-Prey interaction across different models.}
\label{tab:multi_model_causality}
\resizebox{\columnwidth}{!}{
\begin{tabular}{l l c c c}
\hline
\textbf{Model} & \textbf{Direction} & \textbf{Strength} & \textbf{Pred $\rightarrow$ Prey} & \textbf{Prey $\rightarrow$ Pred} \\
\hline
$DPE$   & Pred $\rightarrow$ Prey & 0.1125 & 0.1700 & 0.2825 \\
$ETC_P$ & Pred $\rightarrow$ Prey & 1.0000 & 0.0000 & 1.0000 \\
$ETC_E$ & Pred $\rightarrow$ Prey & 0.0778 & 0.8000 & 0.7222 \\
$LZ_P$  & Pred $\rightarrow$ Prey & 5.0000 & 2.0000 & 7.0000 \\
\hline
\end{tabular}
}
\end{table}

The $DPE$, $ETC_E$, and $LZ_P$ correctly identify the dominant causal direction from predator to prey. After removal of the first nine transient pairs and analyzing the remaining 62 pairs of samples, the weighted entropy calculated by $DPE$ for the $Predator\rightarrow Prey$ direction is lower than that of the reverse $Prey\rightarrow Predator$ seen in table \ref{tab:multi_model_causality}, indicating higher causal certainty in the desired direction ($Predator\rightarrow Prey$). All models, detects the strength of causation from Predator to Prey as higher.

\section{Discussion}\label{sec:discussion}
 Table~\ref{tab:model_comparison} provides a summary of model reliability across experimental settings rather than reporting exact performance values. A framework is considered reliable if it achieves an average accuracy of at least $80\%$ across trials. Thus, the table serves as a binary indicator of consistent performance under each experimental condition. This representation highlights the domains in which each method can be expected to perform robustly, while avoiding overinterpretation of marginal numerical differences.

\begin{table}[!ht]
\centering
\caption{Summary of reliability across experiments. A check mark ($\checkmark$) indicates that the framework achieves an average accuracy of $\geq 80\%$ across trials for the corresponding experiment, while ($\times$) indicates that it does not. The table reflects reliability, not exact performance values.}
\label{tab:model_comparison}
\setlength{\tabcolsep}{2.5pt} 
\resizebox{\columnwidth}{!}{
\begin{tabular}{lccccc}
\toprule
\textbf{Experiments} & \textbf{$DPE$} & \textbf{$ETC_E$} & \textbf{$ETC_P$} & \textbf{$LZ_P$} \\
\midrule
Delayed Bit-flip          & $\checkmark$ & $\times$ & $\times$ & $\checkmark$ \\
AR(1) Coupling            & $\checkmark$ & $\times$ & $\times$ & $\checkmark$ \\
1D Skew-tent Maps         & $\checkmark$ & $\times$ & $\times$ & $\times$ \\
Sparse Processes          & $\checkmark$ & $\times$ & $\times$ & $\times$ \\
SARS-CoV-2 (RS vs. CW)    & $\times$ & $\times$ & $\checkmark$ & $\checkmark$ &  \\
Predator-Prey             & $\checkmark$ & $\checkmark$ & $\checkmark$ & $\checkmark$ \\
\bottomrule
\end{tabular}}
\end{table}

\section{Limitations and Future Work}
\label{sec:limitations}
The present study does not explicitly account for the presence of confounding variables. In particular, the proposed framework assumes that the observed asymmetric pattern-level influence arises directly from the candidate cause. A natural extension of this work is to investigate the identification of latent confounders. One possible direction is to detect patterns that act as common triggers in both sequences, suggesting the presence of an unobserved variable influencing $X$ and $Y$ simultaneously.

Another promising direction is to extend the framework toward a counterfactual formulation of causal discovery. Specifically, we aim to investigate whether the removal of a particular trigger pattern from the assumed cause sequence alters the observed transitions in the effect sequence. Such a formulation would allow us to assess causal influence through pattern-level interventions, thereby strengthening the interpretability and robustness of the method.

For the proposed DPE method, in both the unidirectional coupling experiments and the 1D skew tent map experiments, the case corresponding to $\eta = 0$ or $\phi = 0$ represents true independence between the systems. However, the proposed method as well as all baseline methods considered for comparison fails to correctly identify this regime as independent, instead detecting a spurious directional influence. This limitation indicates that the current decision rule for inferring independence requires further investigation. In particular, the criterion used to distinguish weak causal influence from genuine independence may not be sufficiently stringent in finite-sample settings or in deterministic chaotic systems. To address this issue, future work should incorporate surrogate data analysis and formal statistical significance testing to evaluate the strength and reliability of the inferred causal direction. Establishing appropriate null models and hypothesis-testing procedures would help differentiate true causal structure from artifacts arising due to finite data length, shared dynamics, or intrinsic complexity of the underlying maps.

\section{Conclusion}\label{sec:conclusion}
In this work, we introduced a novel \emph{Dictionary Based Pattern Entropy ($DPE$)} framework for causal discovery from temporal observational data. Beyond inferring causal direction, the proposed approach identifies interpretable sub pattern (algorithmic units) that influence the change in effect variable. This dual capability distinguishes $DPE$ from existing algorithmic information theoretic approaches. Conceptually, the framework operates at the intersection of \emph{Algorithmic Information Theory} (AIT) and \emph{Shannon Information Theory}. From the algorithmic standpoint, causation is interpreted as the emergence of compact, rule based patterns in the candidate cause that systematically constrain the effect. A dictionary of recurring patterns is constructed to capture these algorithmic structures. From the information theoretic perspective, entropy based measures quantify the determinism associated with each pattern, thereby linking structured rule extraction with stochastic variability in observational data. The comparative results summarized in Table~\ref{tab:model_comparison} clarify the relative standing of $DPE$ against competing AIT-based methods ($ETC_E$, $ETC_P$, and $LZ_P$). Using an average accuracy threshold of $80\%$ to determine reliable applicability, $DPE$ demonstrates consistent performance across synthetic dynamical systems (Delayed Bit-flip, AR(1) Coupling, 1D Skew-tent Maps) and structured sparse processes, outperforming or matching competing approaches in most controlled settings. Notably, $DPE$ is the only method that maintains reliability across all synthetic experiments considered. In ecological data, performance across methods is more comparable, whereas in the SARS-CoV-2 mutation analysis, alternative pattern based approaches show competitive advantages over $DPE$. Overall, the results indicate that $DPE$ offers robust and stable causal detection in systems characterized by structured pattern transmission and deterministic influence, while remaining competitive in real-world scenarios. These findings position $DPE$ as a general and interpretable framework for causal discovery, particularly well suited for dynamical systems where causation manifests through identifiable algorithmic sub-patterns rather than purely global complexity measures.
\subsubsection*{Code Availability}
The source code for all experiments, including the implementation of the proposed $DPE$ framework, is publicly available at \href{https://github.com/i-to-the-power-i/dpe-causal-discovery}{https://github.com/i-to-the-power-i/dpe-causal-discovery}.
\begin{contributions} 
Conceptualization: Harikrishnan N B; methodology: Harikrishnan N B, Aditi Kathpalia, Shubham, Nithin Nagaraj; software: Shubham Bhilare (Final implementation) Harikrishnan N B (initial prototype); validation: Shubham Bhilare, Harikrishnan N B, Aditi Kathpalia, Nithin Nagaraj; formal analysis:  Harikrishnan N B.,  Shubham Bhilare, Aditi Kathpalia, Nithin Nagaraj; investigation: Shubham Bhilare, Harikrishnan  N B, Aditi Kathpalia, Nithin Nagaraj ; writing---original draft preparation: Harikrishnan  N B and Shubham Bhilare; writing---review and editing, Harikrishnan  N B, Shubham Bhilare, Aditi Kathpalia and Nithin Nagaraj; Funding acquisition: Harikrishnan N B. 
\end{contributions}

\begin{acknowledgements} Harikrishnan N. B. gratefully acknowledges the financial support from the Prime Minister's Early Career Research Grant (Project No. ANRF/ECRG/2024/004227/ENS).
 
\end{acknowledgements}

\end{document}